\documentclass{article}

     \PassOptionsToPackage{numbers, compress}{natbib}

\usepackage[preprint]{neurips_2020}

\usepackage[document]{ragged2e}
\usepackage[utf8]{inputenc} 
\usepackage[T1]{fontenc}    
\usepackage{hyperref}       
\usepackage{url}            
\usepackage{booktabs}       
\usepackage{amsfonts}       
\usepackage{nicefrac}       
\usepackage{microtype}      
\usepackage{graphicx}
\usepackage{algorithm,algorithmic}
\usepackage{multicol}
\usepackage{amsmath}
\usepackage{subfigure}
\usepackage{ragged2e}

\title{BI-MAML: Balanced Incremental  Approach for Meta Learning}

\author{%
  Yang Zheng\\
  Department of Electrical \& Computer Engineering\\
  University of Washington\\
     \And
  Jinlin Xiang\\
  Department of Mechanical Engineering\\
  University of Washington\\
      \And
  Kun Su\\
  Department of Electrical \& Computer Engineering\\
  University of Washington\\
   \And
   Eli Shlizerman \\
   Department of Applied Mathematics \\
   Department of Electrical \& Computer Engineering \\
   University of Washington \\
}

\begin{document}

\maketitle

\begin{abstract}
\justifying  
We present a novel Balanced Incremental Model Agnostic Meta Learning system (BI-MAML) for learning multiple tasks. Our method implements a meta-update rule to incrementally adapt its model to new tasks without forgetting old tasks. Such a capability is not possible in current state-of-the-art MAML approaches. These methods effectively adapt to new tasks, however, suffer from 'catastrophic forgetting' phenomena, in which new tasks that are streamed into the model degrade the performance of the model on previously learned tasks. Our system performs the meta-updates with only a few-shots and can successfully accomplish them. Our key idea for achieving this is the design of balanced learning strategy for the baseline model. The strategy sets the baseline model to perform equally well on various tasks and incorporates time efficiency. The balanced learning strategy enables BI-MAML to both outperform other state-of-the-art models in terms of classification accuracy for existing tasks and also accomplish efficient adaption to similar new tasks with less required shots. We evaluate BI-MAML by conducting comparisons on two common benchmark datasets with multiple number of image classification tasks. BI-MAML performance demonstrates advantages in both accuracy and efficiency. 
\end{abstract}

\section{Introduction}
\justifying

Humans can learn new information during their life without forgetting the previous information that they have learned, however, this intrinsic capability is hard to achieve in common deep learning methods due to  catastrophic forgetting~\cite{pfulb2019comprehensive} where learning a new task comes at the price of abandoning old tasks. While one intuitive solution is to train for new and old tasks altogether, this strategy becomes infeasible due to the rapid increase in the amount of data to be considered when the number of new tasks is incremented. To address this challenge, incremental learning approach was proposed. An ideal incremental learning strategy is to learn tasks as they come in and use a reasonable amount of data to learn them without degrading performance of previous tasks. 

In addition to incremental learning, the brain is also generalizing the acquired knowledge and adapts it to new tasks. For example, a JAVA programmer may learn a new programming language much faster than a non-programmer. In other words, humans can learn new knowledge based on former experience with only a few new examples. This ability is called meta-learning which is recently being investigated extensively since could provide generic and flexible learning methodologies independent of the particular tasks. To account for these aspects one common strategy is to learn a general baseline model which can be then quickly adapted to a new incoming task using a few samples.

\begin{figure*}[t]
    \centering
    \includegraphics[width=0.9\linewidth]{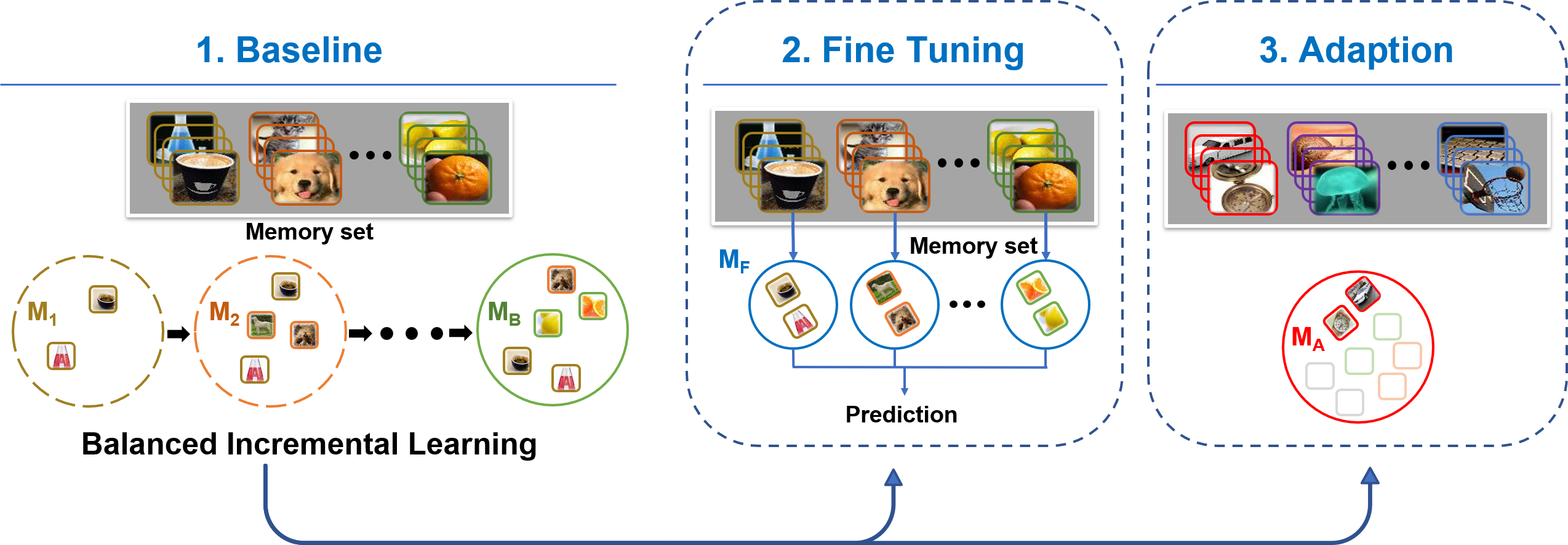}
    \caption{BI-MAML system overview: 1.Baseline: Model $M_1$ evolves to Model $M_B$ by incrementally learning new tasks containing different classes.  2.Fine-Tuning: Model $M_B$ is fine-tuned to Model $M_F$ with specific classes in memory set. 3.Adaption: Model $M_B$ is trained on new task samples. This updates the model to $M_A$ which can solve the new task.}
    \label{fig::overview}
\end{figure*}

While previous work achieve success in one of the aspects of general learning, i.e., incremental learning or adaption, only a few studies have been able to effectively combine these two important properties of learning. In this work we introduce a Balanced Incremental Model Agnostic Meta Learning (\textit{BI-MAML}) approach which combines these two practices of learning. BI-MAML benefits from both incremental and meta learning by learning a baseline model that can generalize and balance both previously learned and new tasks incrementally and efficiently. The BI-MAML system is divided into three phases, namely, ($1$) Baseline, ($2$) Fine Tuning, ($3$) Adaption (See Fig.\ref{fig::overview}). During the Baseline (Fig.\ref{fig::overview}-left), phase the model incrementally learns a sequence of tasks without forgetting previously learned ones and generates a baseline model $M_B$ and a constant-size memory set which store samples from all trained tasks. Then during Fine Tuning (Fig.\ref{fig::overview}-middle) the model incremental testing procedures are implemented where the model is required to perform tests on all learned tasks. In this phase, the baseline model is fine tuned into a task-oriented sub-model $M_F$ using the samples stored in the memory. Summarization of the predictions from $M_B$ and $M_F$ generates enhanced performance and outperforms state-of-the-art methods on several benchmark datasets. In the third phase of Adaption,  (Fig.\ref{fig::overview}-right) the baseline model $M_B$ is adapted to $M_A$ to be able to learn a new task with only a few samples available (few-shot learning). 
In summary, our main contributions with BI-MAML are as follows: (i) we introduce BI-MAML which benefits from both incremental and meta learning. (ii) BI-MAML learns the base model significantly more efficiently than existing meta learning methods. (iii) BI-MAML obtains higher accuracy in incremental classification problems. (iv) BI-MAML achieves state-of-the-art performance on adaption (meta testing). We demonstrate our results on MNIST~\cite{lecun1998gradient} and CIFAR-100~\cite{krizhevsky2009learning} benchmark datasets.

\section{Related Work}
Our model incrementally incorporates new tasks by learning a baseline model reconciling information from both previously learned and new tasks. This generalization enables our model to quickly adapt to a new task with few shots (meta learning) or to learn a new task without losing information on previously learned tasks (incremental learning). Thus we organize the discussion of related work into two major parts: (i) Meta Learning and (ii) Incremental Learning.

\subsection{Meta Learning}
Meta learning is the general approach for specifying flexible and generic rules for a system to learn new tasks. A particular branch of meta learning specifies a meta-updater which will be the general updater for all tasks. An LSTM network was shown to be capable to learn such a general optimizer for all trained tasks~\cite{ravi2016optimization}. Since the update rule can be applied to any model, the method does not require a specific model for each task. However, the generality comes with the cost. Such methods require an additional network for the updater and turn out to be inefficient in terms of determination of meta parameters. In order to improve the learning efficiency of meta parameters without introducing extra networks, alternative methods propose to learn a single general model for all of the considered tasks. Model-Agnostic Meta-Learning (MAML)~\cite{finn2017model} learns this general model by computing a high-order gradient from tracing back an inner loop path along each task. This high-order derivative could be computationally expensive and several methods were proposed to enhance the computational efficiency. For example, First-order MAML (FOMAML)~\cite{nichol2018first} achieves an improvement in computational efficiency by ignoring high-order terms and taking into account the gradient at the last step of the inner loop path. Concurrently, Reptile~\cite{nichol2018reptile} proposed to sum the gradients of all of the steps on the one hand to avoid the high-order derivatives and on the other to keep all the information along the inner loop path. All of the aforementioned model-based methods use the accumulated inner loop gradients to update the outer loop to eventually reach a general model. While these methods are designed during the adaption stage, to adapt to new tasks with a few shots and achieve state-of-art accuracy, the adapted model typically overrides previous knowledge and suffers from catastrophic forgetting. It is therefore advantageous to develop a meta-updater that can remember the previously learned tasks without losing the quick adaption property in meta learning. This is our motivation to develop BI-MAML which will be able to address this kind of meta learning in an incremental way.

\subsection{Incremental learning}

Incremental learning (IL) intents to continuously learn new tasks without forgetting. Promising approaches within IL are knowledge retention-based methods inspired from biological system~\cite{hebb2005organization}. This set of methods retains previous knowledge by imposing constraints on the updating rule of weights according to various strategies recently proposed~\cite{li2017learning, aljundi2018memory, kirkpatrick2017overcoming, zenke2017continual, rebuffi2017icarl, castro2018end, chaudhry2018riemannian}. To prevent the network from forgetting previously learned tasks, it was proposed in~\cite{li2017learning, castro2018end} to apply a distillation loss which considers two networks, a current network and a former network. It then encourages predictions by the former network and the current network to be similar. Furthermore, the two networks jointly learn the representation network and the classification network which could result in inefficiency~\cite{castro2018end}. To enhance the representation of features, distillation-based representation methods and prototype rehearsal were proposed in ~\cite{rebuffi2017icarl}. The network is able to classify the testing samples when it is combined with nearest-mean-of-exemplars rule. Alternatively to distillation, it has been proposed to set constraints to the weights parameter directly ~\cite{kirkpatrick2017overcoming,zenke2017continual}. Specifically, quadratic penalty is applied to parameters to selectively slow down learning which are important for particular tasks and to avoid their forgetting. The importance of each weight is determined by a Fisher information metric.

While the aforementioned methods achieve state-of-the-art performance, time efficiency and model capacity are two main drawbacks to their general applicability. These approaches tend to be time consuming since they require a smaller learning rate to keep the new learned model similar to previous one, however, smaller learning rate prevents the model to rapidly adapt to new tasks which results in time inefficiency. They also tend to be with finite capacity since as the same network is used over time, it  eventually reaches its limits and has to make the trade-offs between acquiring new tasks and holding the information on the previous tasks. In order to improve these deficiency, meta learning-based methods were proposed recently for incremental learning~\cite{riemer2018learning,javed2019meta,rajasegaran2020itaml}. This set of methods learns a model that generalizes both previous and new tasks. Such a general model can be then either adapted, using samples stored in memory, to perform any learned tasks, or to be further trained to learn new tasks. The approach of a general model hence resolves the time efficiency problem of retention-base methods and in terms of model capacity, meta learning-based methods tend to linearly increase in the amount of total parameters (vs nonlinear increase in retention-based models).

The general model can be obtained by solving a minimax problem, where minimization of the chance of future interference and maximization of the transfer of knowledge are mutually optimized~\cite{riemer2018learning}. The iTAML approach~\cite{rajasegaran2020itaml} proposes to uses a meta-update rule to learn a general baseline model for all tasks. As data comes in, it will identify the task and adapt to it in a few update. However, learning a baseline model does not guarantee balanced performance on incremental testing.Without adaption, the baseline model will deteriorate in performance as it acquires new tasks incrementally. BI-MAML, similar to iTAML, approach uses a meta-updater. However, BI-MAML learns the tasks incrementally in a balanced way. It succeeds in it by dividing the network into two parts: Base Feature Extractor and Classifier layers, instead of applying the same training strategy to the entire network. The two parts are being trained in a different way: Base Feature Extractor is trained using knowledge retention-based methods while the Classifier part is trained using a meta-learning-based method. This separate training enables BI-MAML baseline model to achieve balanced and enhanced performance over all tasks.

In addition to those aforementioned branches, IL can also be implemented with several methods, e.g. parameter-efficient model expansion ~\cite{rusu2016progressive, lopez2017gradient, chen2015net2net, chaudhry2018efficient, ren2019incremental}, classifier bias correction ~\cite{wu2019large}, random path selection ~\cite{rajasegaranadaptive, rajasegaran2019random} or dual model consolidation ~\cite{zhang2020class}. 



\begin{figure*}[t]
    \centering
    \includegraphics[width=0.7\linewidth]{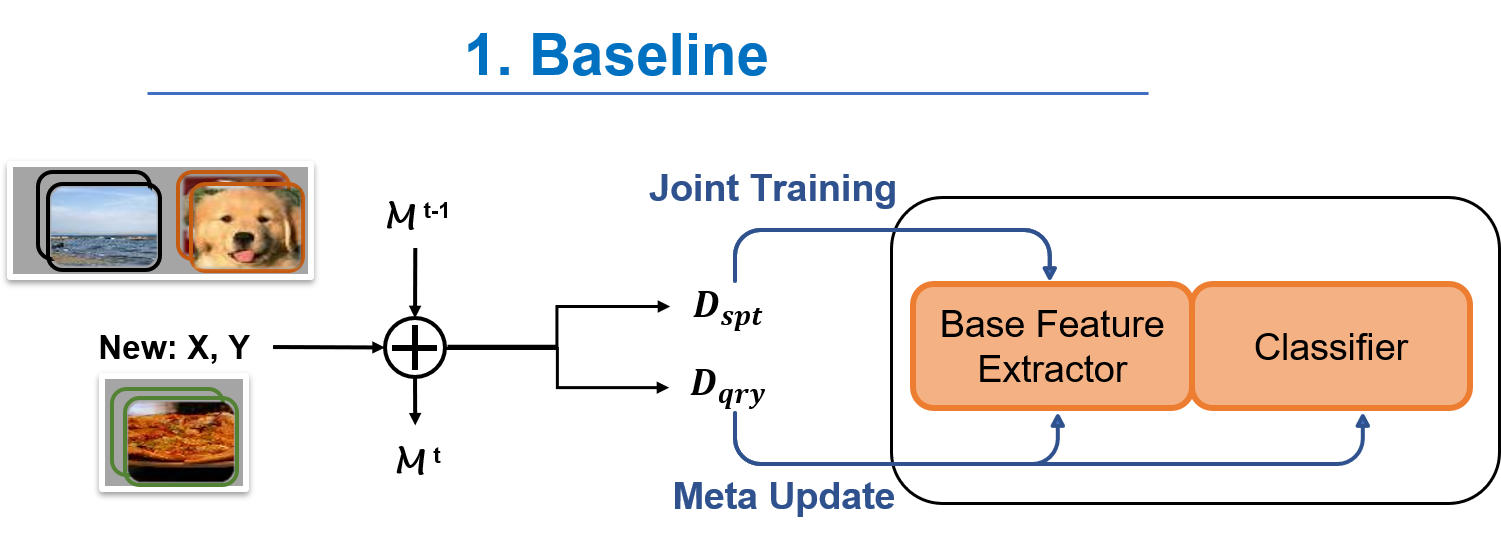}
    \caption{Incremental learning using a meta updater: The training set at task $t$ consisting of $\mathcal{M}^{t-1}$ and a new task $(X^t,Y^t)$ is split into $\mathcal{D}_{spt}$ and $\mathcal{D}_{qry}$ used to train separate blocks, Base Feature Extractor and Classifier respectively. $\mathcal{D}_{spt}$ is used to train the Base Feature Extractor while $\mathcal{D}_{qry}$ uses a meta updater to train the entire network. After training, $\mathcal{M}^{t-1}$ and part of $(X^t,Y^t)$ are reorganized as  $\mathcal{M}^{t}$.}
    \label{fig::pahse1}
\end{figure*}

\section{Methods}
The key component of BI-MAML is the ability to generalize the baseline model, the first step in BI-MAML (and MAML) approach, to perform well on both new and previous tasks. Retaining the knowledge of previous tasks in an incremental setup requires to go back and reconcile the new task information with previous tasks. This corresponds to the need to propagate gradients backward over a chain of tasks that increases in size as the incremental learner continues to acquire them. Such a process could be time consuming and typically slow to converge. BI-MAML addresses this problem by proposing a meta training strategy. Our baseline model is based on gradient approximation which computes the gradients efficiently~\cite{nichol2018reptile}. To overcome the inaccuracies that can result from approximating the gradients, our meta-adapter computes the Focal loss~\cite{lin2017focal} over all tasks. With these two components, our baseline model is able to achieve a \textit{balanced} performance on all learned tasks. In particular, the performance does not deteriorate for tasks that were acquired earlier on in the learning process and at the same time the tasks are acquired with greater efficiency. 

The balanced baseline model can undergo subsequent steps in the standard MAML methodology, i.e., specialization of the baseline model (Fine-tuning) or few-shot learning of new tasks (Adaption), as illustrated in Fig.\ref{fig::overview}. Since BI-MAML will use the balanced baseline model with broader information on all the tasks for these steps, they are expected to achieve enhancement in efficiency and performance. We describe in detail how each step of the MAML process is implemented in the BI-MAML system.


\subsection{Balanced Incremental Learning for the Baseline model.}
During baseline learning, our model implements an increment learning of a sequence of tasks $\mathcal{T} \supseteq \mathcal{T}^1,\mathcal{T}^2,\ldots,\mathcal{T}^n$. In the image classification problem, each task $\mathcal{T}^i$ consists of $k$ classes. Samples of $j$th class in $i$th task are represented as $(X^i_j, Y^i_j)$ where $X$ and $Y$ are the training data and targets, respectively. The memory set $\mathcal{M}$ contains samples from all learned tasks and is used for training the baseline model. 


The baseline model consists of two parts, base feature extractor and classifier, as shown in Fig.\ref{fig::pahse1}. Firstly, we combine training samples ($X^t$, $Y^t$) from new task $\mathcal{T}^t$ and old samples ($X^{1:t-1}$,  $Y^{1:t-1}$) from last memory set $\mathcal{M}^{t-1}$ as current training set. Then we divide current training set into two parts, support set, $\mathcal{D}_{spt}$ and query set, $\mathcal{D}_{qry}$. We use $\mathcal{D}_{spt}$ to train the base feature extractor which is able to separate each class by learning low-level features. In the meanwhile, we freeze the classifies layer to force the network to come up with a good representation for low-level features, instead of applying higher-level features combination from the last several fully connected layers. We also force the tasks in the $\mathcal{D}_{spt}$ to be distributed evenly and find it preserves balanced performance.
\begin{figure*}[ht]
    \centering
    \includegraphics[scale = 0.3]{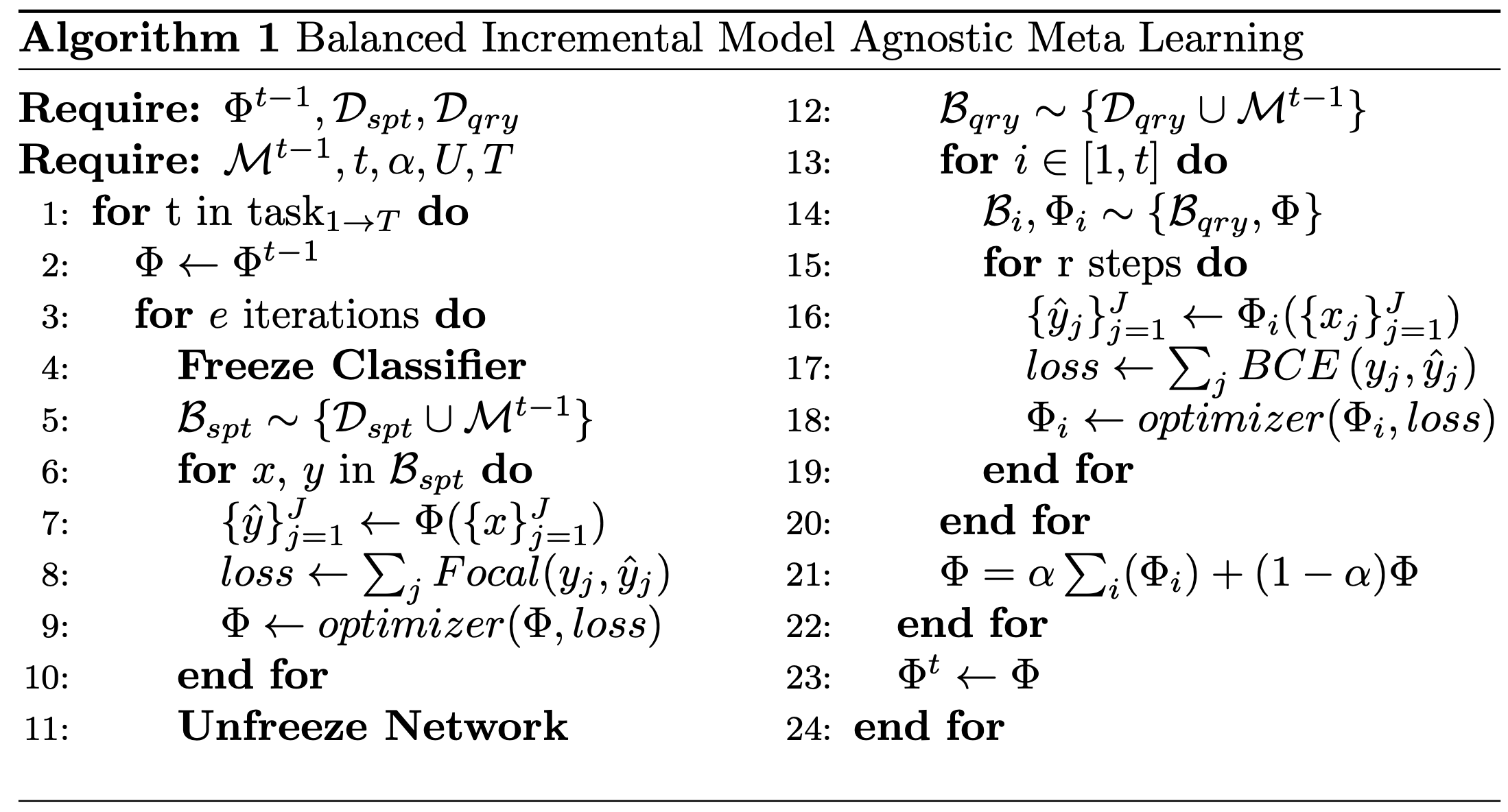}
\end{figure*}
Then we unfreeze the classifier and use meta learning-based strategy to train the entire network using $\mathcal{D}_{qry}$. 
Given features learned from base feature extractor, the classifier aims to learn distinct high-level features for each class. Specifically, we divide the meta learning-based training into two parts, inner updates and outer updates, see Alg.1 for more details.
For simplicity, we use $\Phi$ to represent all parameters in the model. During one iteration, the model will update with each task $i$ from the same initial state $\Phi$ using $\mathcal{B}_i$ which are samples from task $i$ in $\mathcal{D}_{qry}$. Each inner path will arrive at a task-oriented optimal $\Phi_i$. The final update direction in the outer updates is based on the average of all $\Phi_i$. While the $\mathcal{D}_{qry}$ is not always balanced as the new task will usually have more data than the old tasks in the memory, each inner path has the same contribution to the final outer update of the model. This prevents a bias on the new task. Utilizing Focal loss~\cite{lin2017focal} further mitigates imbalance performance since the Focal loss will have more impact on the poor-learned tasks and force the model to achieve a balanced performance. In addition, we apply the idea of knowledge distillation by adding the old task parameter $\Phi^{t-1}$ to the new parameter $\Phi^t$ with a adaptive scale $\alpha$.

After the task is learned, $\mathcal{M}^{t-1}$ will be updated to $\mathcal{M}^{t}$ to include several new samples from $\mathcal{T}^t$. With this training strategy, BI-MAML can efficiently reach state-of-the-art classification accuracy performance however more efficiently.




\begin{figure*}[ht]
    \centering
    \includegraphics[width=0.8\linewidth]{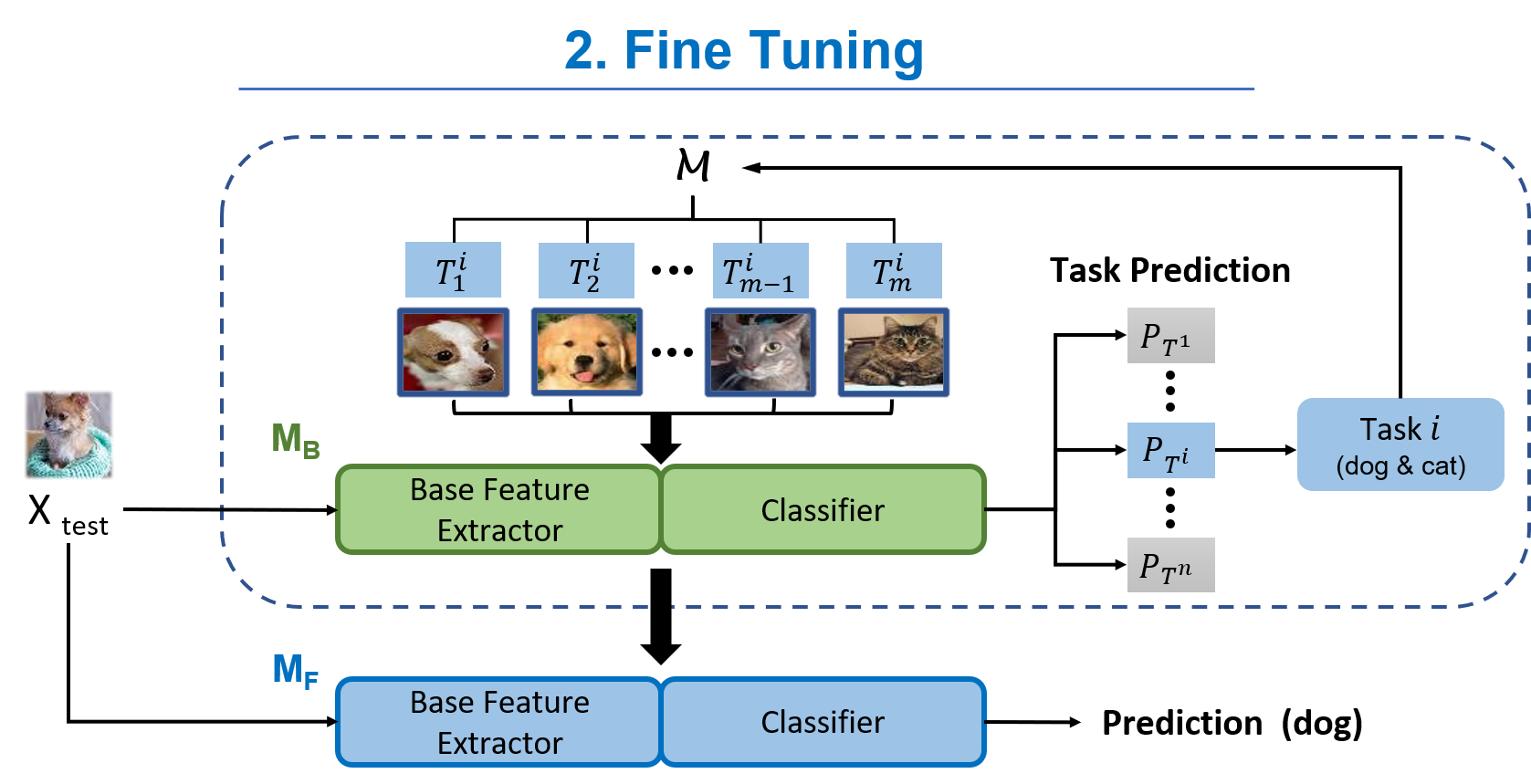}
    \caption{Fine-Tuning: Baseline model $M_B$ first makes task-level prediction of $x_{test}$. Samples of predicted task $T^i$ (dog \& cat) in memory set $\mathcal{M}$ fine tunes $M_B$ to $M_F$. Combining class-level predictions from $M_B$ and $M_F$ is obtained as final prediction(dog).}
    \label{fig::phase2}
    
\end{figure*}
\begin{figure*}[ht]
    \centering
    \includegraphics[scale = 0.35]{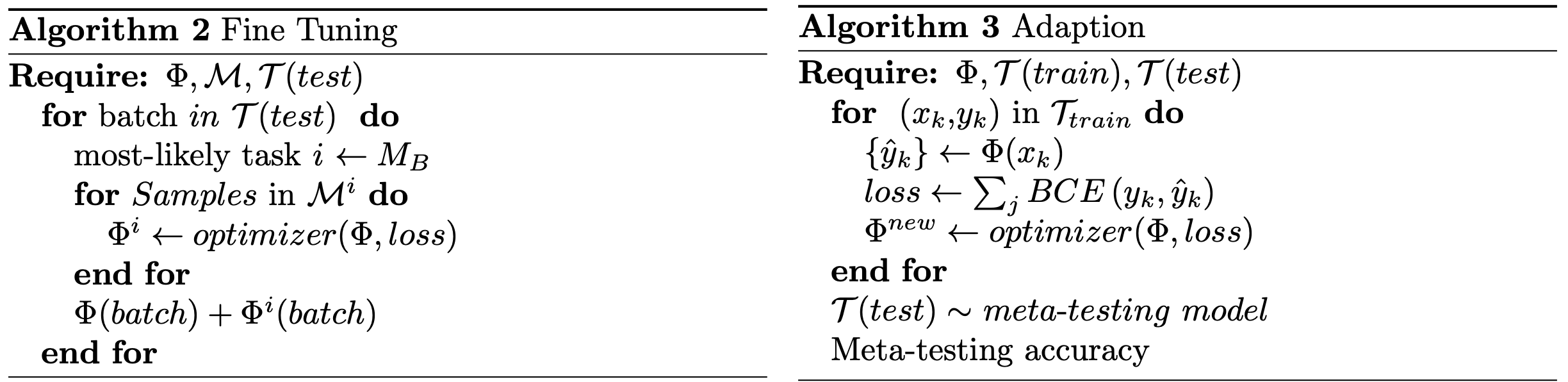}
\end{figure*}

\subsection{Fine tuning}

One important indicator of evaluating an IL algorithm is incremental testing where the model is required to perform on all learned tasks. Even though our baseline model reaches similar classification accuracy as other meta-learning method with much less epochs, its performance is not comparable with other state-of-the-art IL approaches. This limitation cannot be trivially addressed by training for more epochs and requires fine tuning. Thus we apply the fine tuning strategy to enhance our model's performance. This fine tuning can be rapidly done in one-shot to preserve BI-MAML efficiency.

As shown in Fig.~\ref{fig::phase2}, we first use $M_B$ to make a task-level prediction and choose $T_i$ which has the largest probability score as the task label of $x_{test}$. Due to the balanced and relatively high class-level accuracy achieved by the baseline model, this task prediction will have high accuracy and is not biased on any specific task. We fine tune $M_B$ to $M_F$ for a specific $T_i$ using corresponding samples from memory. The final class-level prediction is made by combining two class-level predictions made by $M_B$ and $M_F$, respectively. This fast fine tuning not only enables BI-MAML to obtain state-of-the-art performance on incremental testing, but also to preserve its high efficiency achieved in the baseline model training.


\begin{figure*}[t]
    \centering
    \includegraphics[width=0.8\linewidth]{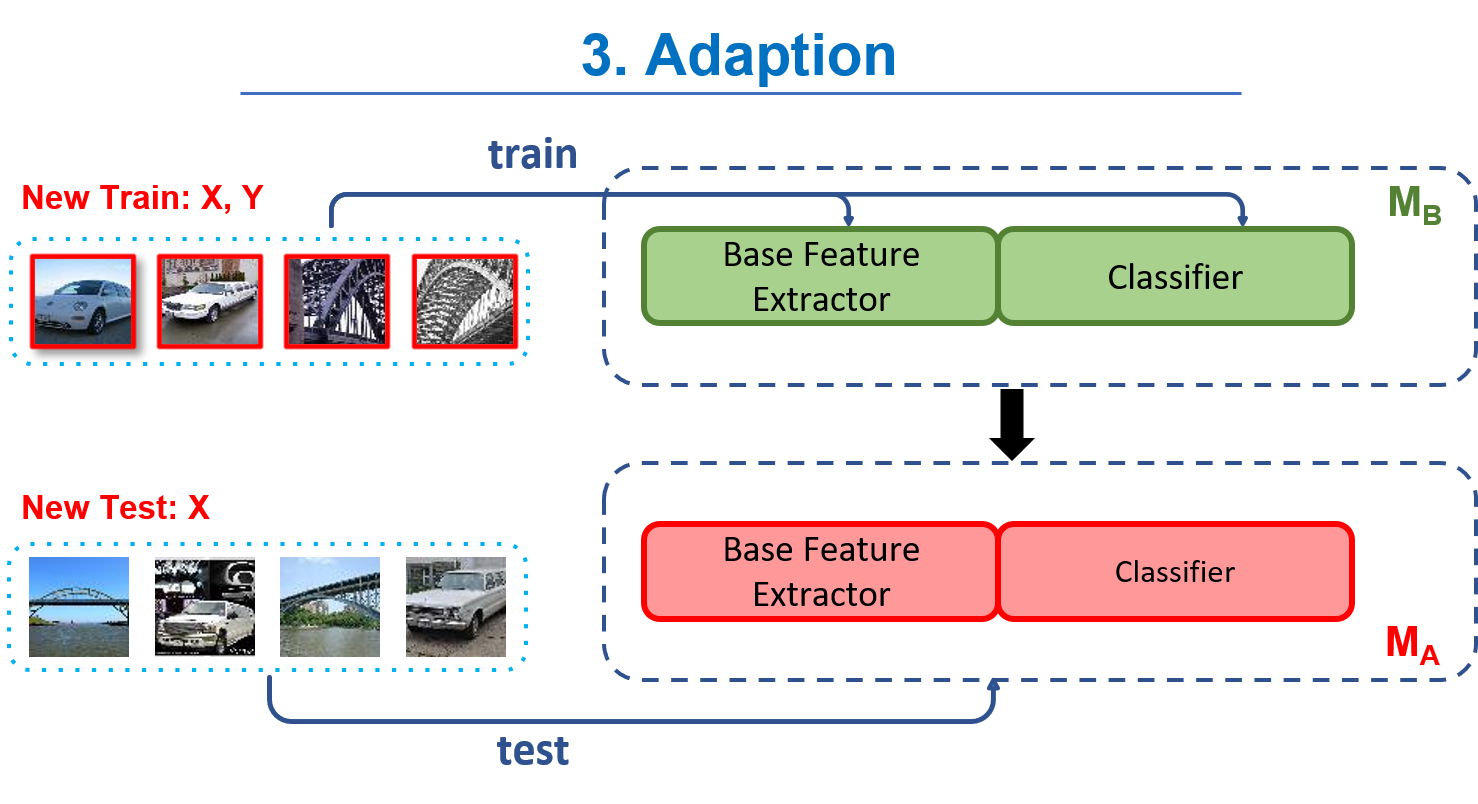}
    \caption{Adaption: $M_B$ is adapted using new train samples $(X, Y)$. The adapted model, $M_A$, is able to solve the new task.}
    \label{fig::phase3}
\end{figure*}

\subsection{Adaption}

The generality of the baseline model is evaluated by meta testing where the model is required to adapt to a new task in few shots using a limited amount of samples. It turns out that balancing and high classification accuracy in each task are two essential properties for a generic model. BI-MAML incorporates these two properties in the baseline model. Our baseline model has balanced high performance over each task which enables it to quickly adapt to new task. Different from common meta learning methods, BI-MAML can learn the new task without suffering from catastrophic forgetting by conducting knowledge retention approach.

Our meta testing procedure in the adaption phase is shown in Fig.~\ref{fig::phase3}. Initially, the new task is divided into two parts, training and testing sets. In contrast to the fine tuning phase, the memory is no longer included. Since imbalances are not an issue in adaption, we train the entire network without freezing. The trained model $M_A$ is then exclusive to this task and its performance can be validated using the testing set.


\section{Experiments \& Evaluation}

\begin{figure}[t]
    \centering
    \includegraphics[width=1\textwidth]{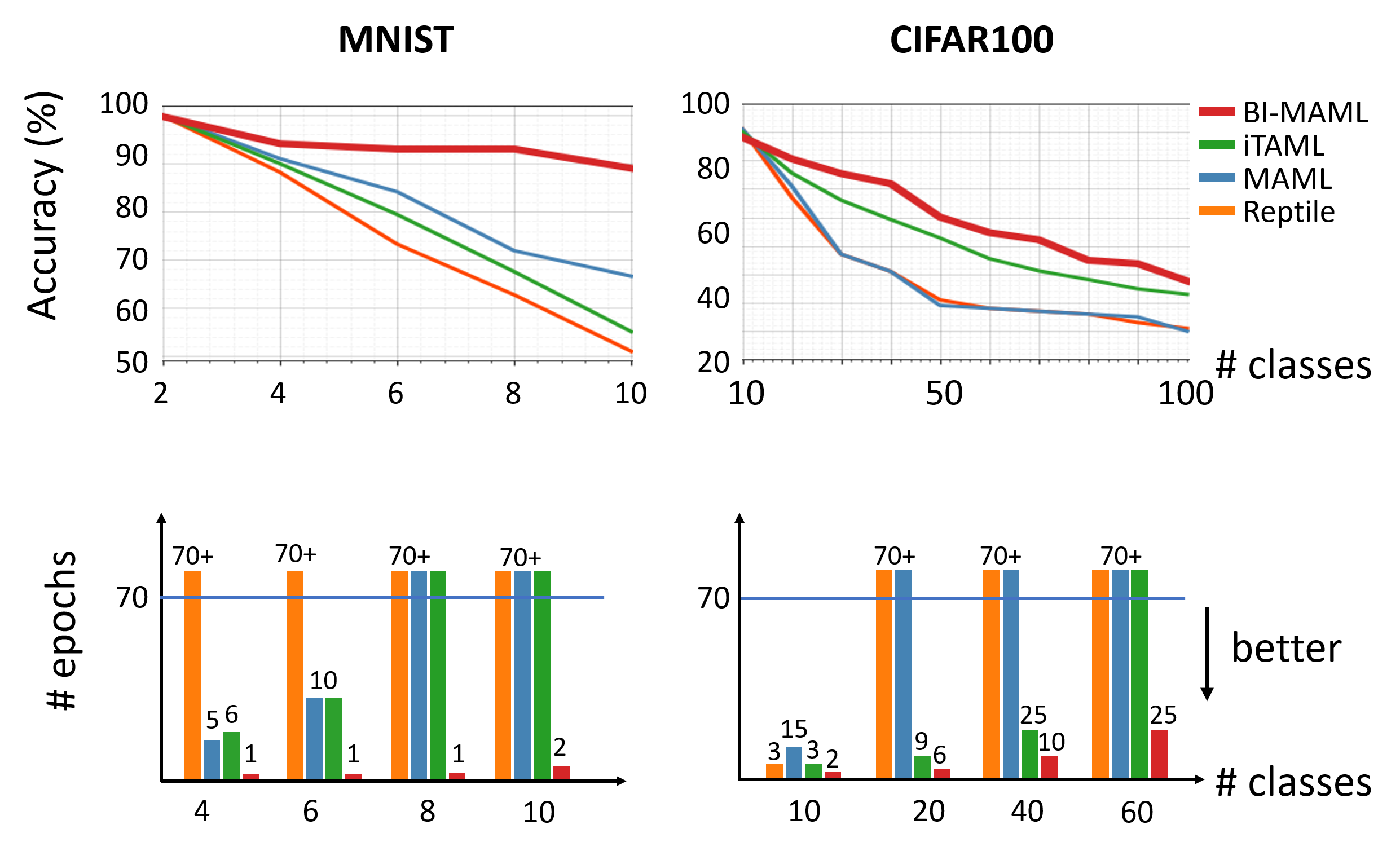}
    \caption{Baseline classification accuracy(top) and efficiency(bottom) on MNIST(left) and CIFAR-100(right) of compared methods(BI-MAML(red), iTAML(green), MAML(blue), Reptile(orange)). Top: Accuracy vs. number of classes(2(left), 10(right) classes per task), Bottom: number of epochs vs. number of classes (90\%(left) and 50\%(right) are used as fixed accuracy).}
    \label{fig:baseline_result}
\end{figure}

\subsection{Implementation Details}
We evaluate our method on several datasets and compare the performance with other methods of incremental learning and meta learning. For small datasets, such as split MNIST and split SVHN~\cite{netzer2011reading}, the baseline model incrementally learns two classes at each time. For larger datasets, like CIFAR-100~\cite{krizhevsky2009learning}, each task consists of 10 classes. In order to test the capability and the robustness of the model, we also test the baseline model with different task sizes in CIFAR-100 (10,20,50 classes for each task). We set the default size of the memory set to 2000. For MNIST, we use a 3 fully connected layers network (400 neurons each). For SVHN and CIFAR-100, we apply RPS net~\cite{rajasegaran2019random} with 1.25M parameters. We use SGD~\cite{lecun2012efficient} as the optimizer for MNIST, and RAdam~\cite{liu2019variance} for SVHN and CIFAR-100. The default number of epochs is 70 and 5 for large and small datasets respectively. We adjust the learning rate by decreasing it by scalar multiplication of 0.2 for each 20 epochs.

\subsection{Results and Comparison}

\textbf{Baseline Model Evaluation:} We compare our baseline model with other meta-learning algorithms~\cite{finn2017model, nichol2018reptile} and iTAML~\cite{rajasegaran2020itaml} in terms of training efficiency and training accuracy. The results in Fig.~\ref{fig:baseline_result} demonstrate the accuracy (top) and efficiency (bottom) on MNIST (left) and CIFAR-100 (right). In terms of accuracy (Fig.~\ref{fig:baseline_result} (top)), BI-MAML outperforms other methods by a large margin on MNIST and the best by a noticeable margin on CIFAR-100. More specifically, our model reaches $88.99\%$ accuracy on MNIST 10 tasks, which is $22.6\%$ higher than iTAML (the second best method). For CIFAR-100, our model achieves on average 5.63\% accuracy increase than the second best method iTAML. 

Fig.~\ref{fig:baseline_result} (bottom) demonstrates the number of epochs for different methods to reach a fixed accuracy ($90\%$ for MNIST and $50\%$ for CIFAR-100). We assume $70$ epochs as a bar and claim this method will not reach this accuracy when it reaches $70$ epochs and did not reach the desired accuracy. We choose $4, 6, 8, 10$ classes as total testing classes in MNIST and $10, 20, 40, 60$ classes for CIFAR-100 separately. Our results indicate that BI-MAML can surpass all other methods by a great margin which demonstrates the significant efficiency benefit of BI-MAML. Both of these benchmarks confirm that BI-MAML is favourable in both accuracy and efficiency among all tested methods.


\textbf{Incremental Methods Comparison:} In this section, we compare the fine tuned BI-MAML with other popular incremental learning methods in terms of accuracy~\cite{li2017learning, rebuffi2017icarl,rajasegaran2020itaml, rajasegaran2019random, zhang2020class}.

\begin{figure}[t]
    \centering
    \includegraphics[width=0.9\textwidth]{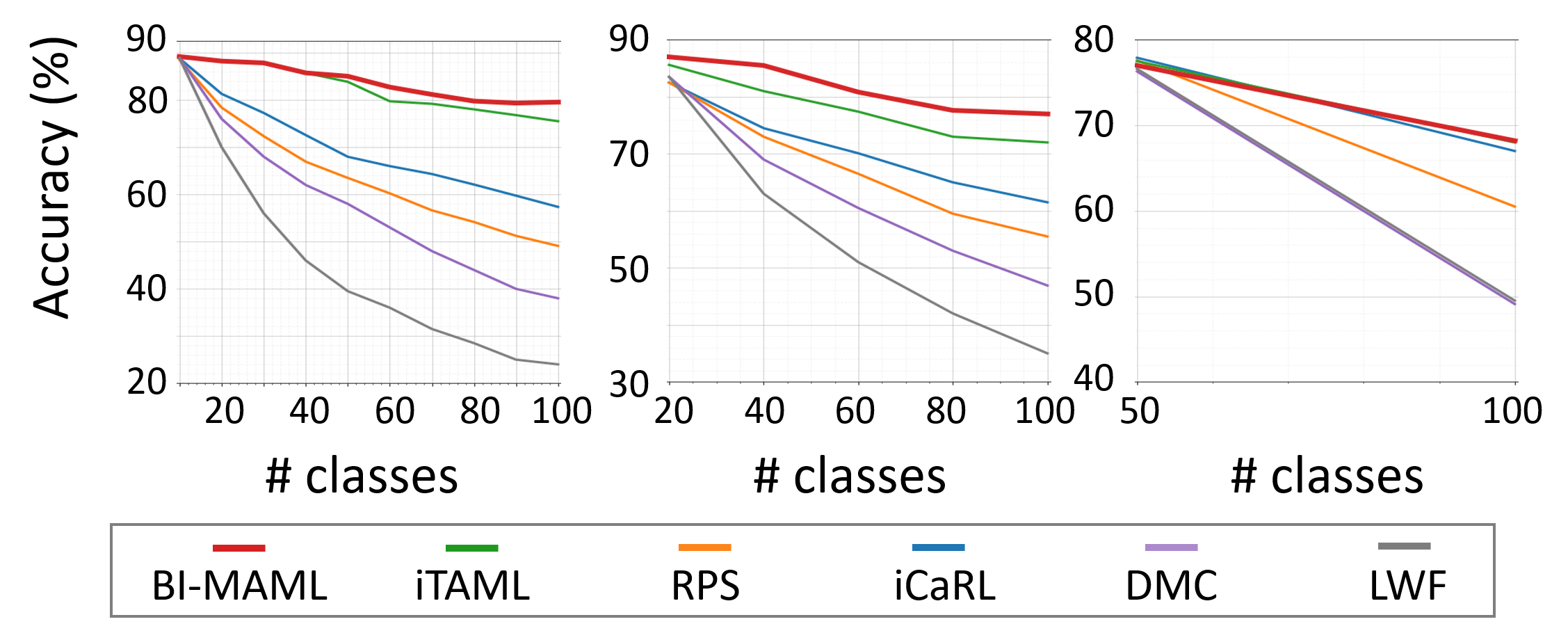}
    \caption{Fine tuning results on CIFAR-100. Left to right are results of learning 10, 20 and 50 classes at each time, respectively.}
    \label{fig::fine_tuning_result}
\end{figure}

As shown in Table.~\ref{table::fine_tuning_result} and Fig.~\ref{fig::fine_tuning_result}, BI-MAML achieves state-of-the-art performance in different datasets tested. For MNIST and SVHN (shown in Table.~\ref{table::fine_tuning_result}), BI-MAML achieves classification accuracy of 99.02\% and 98.42\% respectively. We also illustrate the capability of BI-MAML by changing the size of tasks. In Fig.~\ref{fig::fine_tuning_result}, each method learns 10 (left), 20 (mid) and 50 (right) classes at each time, respectively. 
Quantitatively, BI-MAML reaches $79.56\%$, $77.0\%$ and $68.12\%$ for task of size 10, 20, 50, respectively, and is higher than all other methods.

\begin{table}[htbp]
    \centering
        \begin{tabular}{ccccccc}
        \hline
        method & GEM& DGR & RtF & RPS-net & iTAML & \textbf{BI-MAML(ours)} \\
        \hline
        MNIST & 92.2 & 91.24 & 92.16  & 96.16 & 97.75 & \textbf{99.02} \\
        SVHN & 75.61 & -- & -- & 88.91 & 93.97 & \textbf{98.42}\\
        \hline
        \end{tabular}
    \caption{Fine-tuning comparison on MNIST and SVHN datasets. }
    \label{table::fine_tuning_result}
\end{table}

\textbf{Meta Test (Adaption):} In the meta testing, we test our BI-MAML adaptability to the new task after it is trained. We conduct experiments on MNIST, SVHN and CIFAR-100 and show our results in Table.~\ref{table::adaption_result}. "2w-1s" in the table means two classes are in each task and only one adaption is used. As shown, in these experiments BI-MAML obtains similar accuracy as common meta learning methods. 

\begin{table}[ht]
    \centering
    \begin{tabular}{ccccc}
    \hline
         & MNIST & SVHN & \multicolumn{2}{c}{CIFAR-100}  \\
         & 2w-1s & 2w-1s & 5w-1s & 5w-5s \\
         \hline
        MAML~\cite{finn2017model} & 99.81 & 86.6 & 56.8 & 70.80\\
        \textbf{BI-MAML(ours)} & 98.40 & 85.29 & 55.40 & 71.81\\
        \hline
    \end{tabular}
    \caption{Adaption performance comparison on MNIST, SVHN and CIFAR-100 datasets. 2w-1s means 2-way-1-shot.}
    \label{table::adaption_result}
\end{table}





\section{Conclusion}
Incremental learning aims to overcome catastrophic forgetting while the goal of Meta Learning is to provide general learning rules. BI-MAML combines the two with a balanced training approach for the baseline model. BI-MAML results show significant efficiency leap and accuracy improvement across several datasets: MNIST, SVHN and CIFAR-100.




\bibliographystyle{ieeetr}

\end{document}